\documentclass[final]{cvpr}

\usepackage{amsmath,amsfonts,bm}

\def\eqref#1{equation~\ref{#1}}

\def\1{\bm{1}}

\DeclareMathAlphabet{\mathsfit}{\encodingdefault}{\sfdefault}{m}{sl}
\SetMathAlphabet{\mathsfit}{bold}{\encodingdefault}{\sfdefault}{bx}{n}

\usepackage{times}
\usepackage{epsfig}
\usepackage{graphicx}
\usepackage{amsmath}
\usepackage{amssymb}

\usepackage{url}
\usepackage{subcaption}
\usepackage{graphicx}
\usepackage{multirow}
\usepackage{booktabs}
\usepackage{color}
\usepackage{xcolor}
\usepackage{wrapfig}
\usepackage{adjustbox}
\usepackage[pagebackref=true,breaklinks=true,colorlinks,bookmarks=false]{hyperref}

\def\cvprPaperID{6} 
\def\confYear{CVPR 2021}

\begin{document}

\title{Renofeation: A Simple Transfer Learning Method\\ for Improved Adversarial Robustness}

\author{Ting-Wu Chi$\text{n}^1$, Cha Zhan$\text{g}^2$, Diana Marculesc$\text{u}^{13}$\\
Carnegie Mellon Universit$\text{y}^1$, Microsoft Cloud and A$\text{I}^2$, The University of Texas at Austi$\text{n}^3$\\
{\tt\small tingwuc@andrew.cmu.edu, chazhang@mirosoft.com, dianam@utexas.edu}}

\maketitle

\begin{abstract}
Fine-tuning through knowledge transfer from a pre-trained model on a large-scale dataset is a widely spread approach to effectively build models on small-scale datasets. In this work, we show that a recent adversarial attack designed for transfer learning via re-training the last linear layer can successfully deceive models trained with transfer learning via end-to-end fine-tuning. This raises security concerns for many industrial applications. In contrast, models trained with random initialization without transfer are much more robust to such attacks, although these models often exhibit much lower accuracy. To this end, we propose \emph{noisy feature distillation}, a new transfer learning method that trains a network from random initialization while achieving clean-data performance competitive with fine-tuning. Code available at \url{https://github.com/cmu-enyac/Renofeation}. 
\end{abstract}

\section{Introduction}
\label{intro}
Transfer learning is an important approach that enables training deep neural networks faster and with relatively less data than training from scratch without any prior knowledge. Specifically, we consider the setting where we want to maximize the performance on a target task assuming the availability of a pre-trained model trained on a source task. This setting has various applications and has led to state-of-the-art performance in several image classification tasks~\cite{Cui_2018_CVPR}. Moreover, this setting is also considered in industry in the form of machine-as-a-service, such as Google's Cloud AutoML~\cite{gautoml} and Microsoft's Custom Vision service~\cite{azure} where users can upload custom data to fine-tune a pre-trained model. We refer to this setting as transfer learning throughout this paper.

Transfer learning for ConvNets has received great attention due to its effectiveness in achieving high accuracy. It has been shown~\cite{simonyan2014very} that the pre-trained model trained on a large-scale dataset (such as ImageNet) acts as an effective feature extractor that supersedes hand-crafted feature extractors. Subsequent work~\cite{yosinski2014transferable,donahue2014decaf} has found that inheriting the pre-trained weights and starting learning from there (often referred to as ``fine-tuning") can result in even larger performance improvements. Fine-tuning has then been adopted in various tasks to achieve state-of-the-art results. Besides fine-tuning, several prior methods have relied on fine-tuning with an explicit regularization loss to further enhance the performance of transfer learning~\cite{xuhong2018explicit,li2018delta}. While prior art has demonstrated that fine-tuning might not necessarily outperform training from random initialization for some tasks, such as classifying medical images~\cite{NIPS2019_8596} and object detection and semantic segmentation with sufficient training data~\cite{he2019rethinking}, it is important to note that fine-tuning is the state-of-the-art method for small and visually similar datasets such as the Caltech-UCSD Bird 200 datasets~\cite{WelinderEtal2010}.

Very recently it has been demonstrated~\cite{Rezaei2020A} that models transferred by re-learning the last linear layer are vulnerable to adversarial examples crafted solely based on the pre-trained model. In other words, an adversary can attack a pre-trained model available on open repositories, \textit{e.g.}, TorchVision, and use the adversarial image to deceive the transferred models. In this paper, we show that such an attack can also deceive models transferred with end-to-end fine-tuning. This finding raises security concerns for the widely-adopted fine-tuning mechanism, which is also used in industrial applications such as Google's AutoML~\cite{gautoml} and Microsoft's Custom Vision~\cite{azure}. In this work, we take a first step toward alleviating this problem. Intuitively, the vulnerability to such an attack stems from the similarity between the pre-trained and the transferred models. However, we find that models transferred with existing fine-tuning methods are similar to the pre-trained ones, which in turn makes them vulnerable to the attack developed by Rezaei \textit{et al.}~\cite{Rezaei2020A}. In contrast, models trained with random initialization are much more robust to such attacks, with the caveat that these models often exhibit much lower accuracy compared to fine-tuning. As an alternative to prior methods, we propose \underline{re}-training with \underline{no}isy \underline{fea}ture distilla\underline{tion} (or \textbf{Renofeation} for short), which achieves clean-data performance similar to fine-tuning and the robustness of training with random initialization. Overall, our contributions are as follows: 

\begin{itemize}
    \item We show that the attack proposed in prior work is suitable not only for transfer learning by re-training the last linear layer, but also for transfer learning with end-to-end fine-tuning, which raises security concerns for the widely-adopted fine-tuning paradigm.
    \item We propose Renofeation, a new transfer learning method that results in competitive clean-data performance compared to fine-tuning with significant better robustness. \textbf{Compared to previous transfer learning methods, ours is the first that argues for ``reinitializing the weights"}.
    \item We conduct extensive experiments on four networks and five datasets with hyper-parameter tuning and ablation studies to empirically demonstrate the effectiveness of the proposed method.
\end{itemize}


\section{Background}
\subsection{Transfer learning}
It is known that deep neural networks trained on large-scale datasets such as ImageNet learn surprisingly transferable features~\cite{simonyan2014very}. That is, one can re-purpose a pre-trained network to other classification tasks by simply learning a linear classifier on top of the features from the penultimate layer. Later, researchers have found that when the entire pre-trained model is optimized with a small learning rate, performance can be even better~\cite{Li2020Rethinking,donahue2014decaf,raghu2019transfusion,he2019rethinking,yosinski2014transferable}, and this scheme is also known as ``fine-tuning". With the desire of not forgetting the useful features learned from the large-scale dataset, explicit regularization was proposed to further improve fine-tuning. Specifically, L2SP~\cite{xuhong2018explicit,lee2019mixout} imposes regularization to avoid weights deviating from the pre-trained weights in a $\ell_2$ sense. Similarly, DELTA~\cite{li2018delta,pmlr-v97-jang19b,wang2020pay} imposes regularization to avoid representations deviating from the pre-trained representations. Besides these transfer learning methods, training from random initialization is often considered as the baseline for transfer learning, which does not leverage the information learned from the pre-trained model.

Transfer learning using extra information or architectural changes have also been investigated in the literature. Ge \textit{et al.}~\cite{ge2017borrowing} developed a method to improve fine-tuning by leveraging additional training data obtained from large-scale datasets. Cui \textit{et al.}~\cite{Cui_2018_CVPR} used Earth Mover's Distance to measure domain similarity between datasets and showed that pre-training on similar domains results in better transfer. Wang \textit{et al.}~\cite{wang2017growing} discovered that increasing the model capacity (wider or deeper) improves the effectiveness of fine-tuning.

\subsection{Adversarial examples}
Adversarial examples~\cite{szegedy2013intriguing} for deep learning models have received growing attention due to their potential impact on machine learning systems. According to different threat models, there are various types of attacks. In a white-box threat model, where the adversary knows all the information regarding a model, fast gradient sign method (FGSM)~\cite{goodfellow2014explaining}, projected gradient descent, and CW~\cite{carlini2017towards} have been shown to be strong attacks. Counteracting these attacks, adversarial training~\cite{madry2017towards} is the dominant approach for robusifying deep networks. On the other hand, there are also methods targeting a black-box threat model where the adversary can only query the model and obtain the probability vector~\cite{liu2016delving,papernot2016transferability,chen2017zoo}. 

In this work, our threat model assumes that the adversary has access to the \textit{model weights} and \textit{model architecture} for the \textit{pre-trained model}. The adversary \textit{does not} have access to the \textit{task-specific transferred model}. This threat model aligns with practical usage of deep learning models where researchers use pre-trained models on large datasets (like ImageNet) and fine-tune them for other tasks. Based on this threat model, prior art~\cite{Rezaei2020A} has proposed an attack that successfully compromises the task-specific transferred models, which raises security concerns for transfer learning. In this work, we find that such an attack not only successfully deceives transfer learning by re-training the last linear layer, but also works for end-to-end fine-tuning. We further propose an algorithm to improve the robustness of the transferred model under this particular threat model. On a different threat model, Shafahi \textit{et al.}~\cite{shafahi2020adversarially} have proposed to improve the adversarial robustness of the transferred model in a white-box setting by transferring to the target model the robust features obtained through adversarial training. While in this work we use feature distillation to improve clean data performance, knowledge distillation has been explored to improve the robustness of the student model by distilling from a robust teacher~\cite{goldblum2019adversarially}.

To craft an adversarial example under our threat model, we adopt an attack from Rezaei \textit{et al.}~\cite{Rezaei2020A}, which optimizes the following objective:
\begin{equation}\label{eq:perturb}
    \begin{split}
        \arg\min_{\delta}~&\lVert f_{K}(x+\delta,\theta_0) -  t \rVert_2^2\\
        s.t.~&\lVert \delta \rVert_{\infty} \leq B,
    \end{split}
\end{equation}
where $f_K$ is the output of the penultimate layer, $t$ is a target vector that is set to a scalar $m$ multiplied by a one-hot vector. $m$ is chosen to be large and $B$ denotes the perturbation budget. The pixel intensity in this formulation is normalized and constrained to $[0,1]$. We optimize equation~\ref{eq:perturb} via projected gradient descent (PGD). Intuitively, the objective is trying to find a small-norm perturbation such that the response of the penultimate layer of the pre-trained model is polarized. Once the perturbation $\delta$ for a specific input image $x$ is found, the perturbed image $x+\delta$ is used to attack a transferred model $\theta$.

\section{Motivation}\label{sec:motivation}
\begin{table*}[h]
\caption{Summary of different transfer learning methods. $\bm{\theta}$ and $\bm{\theta}_0$ denote the weights for the transferred and pre-trained neural network, respectively. $f_l(\cdot,\cdot)$ denotes the output (feature) of the $l^{\text{th}}$ layer.}
\begin{center}
\begin{small}
\begin{sc}
\begin{adjustbox}{max width=1\linewidth}
\begin{tabular}{c|c|c|c|c|c|c}
\toprule
& Re-training & Linear classifier & Fine-tuning & L2SP~\cite{xuhong2018explicit} & DELTA~\cite{li2018delta} & Renofeation (Ours) \\
\midrule
Random init. (Layer) & All & Last & Last & Last & Last & All\\
\hline
Variable (Layer) & All & Last & All & All & All & All\\
\hline
\multirow{3}{*}{Regularization} & & & & & & $\sum_{l=1}^L \|f_l(x,\bm{\theta})-f_l(x,\bm{\theta}_0)\|_2^2$\\
& $\|\bm{\theta}\|_2^2$ & $\|\bm{\theta}\|_2^2$ & $\|\bm{\theta}\|_2^2$ & $\|\bm{\theta}-\bm{\theta}_0\|_2^2$ & $\sum_{l=1}^L \|f_l(x,\bm{\theta})-f_l(x,\bm{\theta}_0)\|_2^2$ & Dropout\\
 & & & & & & Stochastic Weight Averaging\\
\midrule
Transfer Mechanism & N/A & Weights & Weights & Weights & Weights and Features & Features\\
\hline
\multirow{2}{*}{Defense Mechanism} & \multirow{2}{*}{Random init.} & \multirow{2}{*}{N/A} & \multirow{2}{*}{N/A} & \multirow{2}{*}{N/A} & \multirow{2}{*}{N/A} & Random init.\\
 & & & & & & Feature regularization\\
\bottomrule
\end{tabular}\label{table:methods}
\end{adjustbox}
\end{sc}
\end{small}
\end{center}
\end{table*}

We start with the following research question: ``\textit{Can the attack proposed by Rezaei \textit{et al.}~\cite{Rezaei2020A} compromise transfer learning that fine-tunes the entire model?}'' This is unclear as such an attack was originally proposed to deceive a specific transfer learning method, \textit{i.e.}, re-learning the last linear layer. Since end-to-end fine-tuning provides much better performance compared to only learning the last linear layer~\cite{li2018delta}, fine-tuning is a widely adopted method for transfer learning. As a result, it would be less concerning if such an attack only works for re-learning the last linear layer but not end-to-end fine-tuning.

To answer this question, we consider five training methods with five transfer learning datasets. For training methods, we consider \textit{Linear classifier} that only re-learn the last linear layer, \textit{Fine-tuning} that trains all the parameters, \textit{L2SP}~\cite{xuhong2018explicit} that trains all the parameters with weight regularization, \textit{DELTA} that trains all the parameters with representation regularization, and a baseline \textit{Re-training} that trains from random initialization using the target dataset without any transfer learning. We summarize the methods used in Table~\ref{table:methods}. As for datasets, we consider Stanford Dog~\cite{khosla2011novel}, Caltech-UCSD Bird~\cite{WelinderEtal2010}, Stanford Actions~\cite{yao2011human}, MIT Indoor Scenes~\cite{quattoni2009recognizing}, and Flower~\cite{nilsback2008automated}. We proceed by crafting adversarial examples by solving equation~(\ref{eq:perturb}) using PGD. Then, we evaluate the \textit{attack success rate (ASR)}, which is calculated by the conditional probability $P($wrong~with~adversarial-data~$|$~correct~with~clean-data$)$, for each method and dataset combination. To provide context, we also evaluate the top-1 image classification accuracy on clean images for each method and dataset combination.

\begin{table}[t]
\caption{Robustness evaluation for the baseline transfer learning methods for ResNet18. ASR denotes attack success rate, which is computed as $P($wrong~with~adversarial-data~$|$~correct~with~clean-data$)$ (the lower the more robust). Clean denotes the Top-1 accuracy for clean-data.}
\begin{center}
\begin{small}
\begin{sc}
\begin{adjustbox}{max width=1\linewidth}
\begin{tabular}{c|c|c|c|c|c|c}
\toprule
\multicolumn{2}{c|}{} & Dog & Bird & Action & Indoor & Flower\\
\midrule
\multirow{2}{*}{Linear classifier} & clean & 84.22 & 67.02 & 73.64 & 72.54 & 88.52\\
& ASR & 96.06 & 96.47 & 92.49 & 88.95 & 86.40\\
\midrule
\multirow{2}{*}{Fine-tuning} & clean & 81.84 & 77.67 & 77.19 & 75.37 & 95.71\\
& ASR & 89.36 & 50.33 & 73.75 & 54.75 & 14.07\\
\midrule
\multirow{2}{*}{L2SP} & clean & 83.82 & 77.51 & 77.22 & 75.15 & 95.63\\
& ASR & 94.08 & 50.08 & 92.16 & 66.73 & 16.75\\
\midrule
\multirow{2}{*}{DELTA} & clean & \textbf{84.39} & \textbf{78.75} & \textbf{77.69} & \textbf{78.36} & \textbf{95.90}\\
& ASR & 95.65 & 58.83 & 93.51 & 79.71 & 43.65\\
\midrule
\multirow{2}{*}{Re-training} & clean & 70.77 & 69.76 & 51.90 & 59.93 & 87.38\\
& ASR & \textbf{5.99} & \textbf{6.14} & \textbf{5.82} & \textbf{6.73} & \textbf{3.00}\\
\bottomrule
\end{tabular}\label{table:benchmark}
\end{adjustbox}
\end{sc}
\end{small}
\end{center}
\end{table}

As shown in Table~\ref{table:benchmark}, we find that the attack proposed by Rezaei \textit{et al.}~\cite{Rezaei2020A} can deceive models trained with end-to-end fine-tuning with high attack success rate. This raises security concerns for the widely-adopted fine-tuning paradigm. Besides confirming that DELTA is the best transfer learning method among the considered ones, we observe that although the clean data performance for re-training is less than ideal compared to transfer learning, it has low attack success rate. This observation leads us to the following question: ``\textit{Why are models trained with end-to-end fine-tuning vulnerable to such an attack while re-training are robust?}'' 

\begin{figure}[h]
    \centering
    \includegraphics[width=0.48\linewidth]{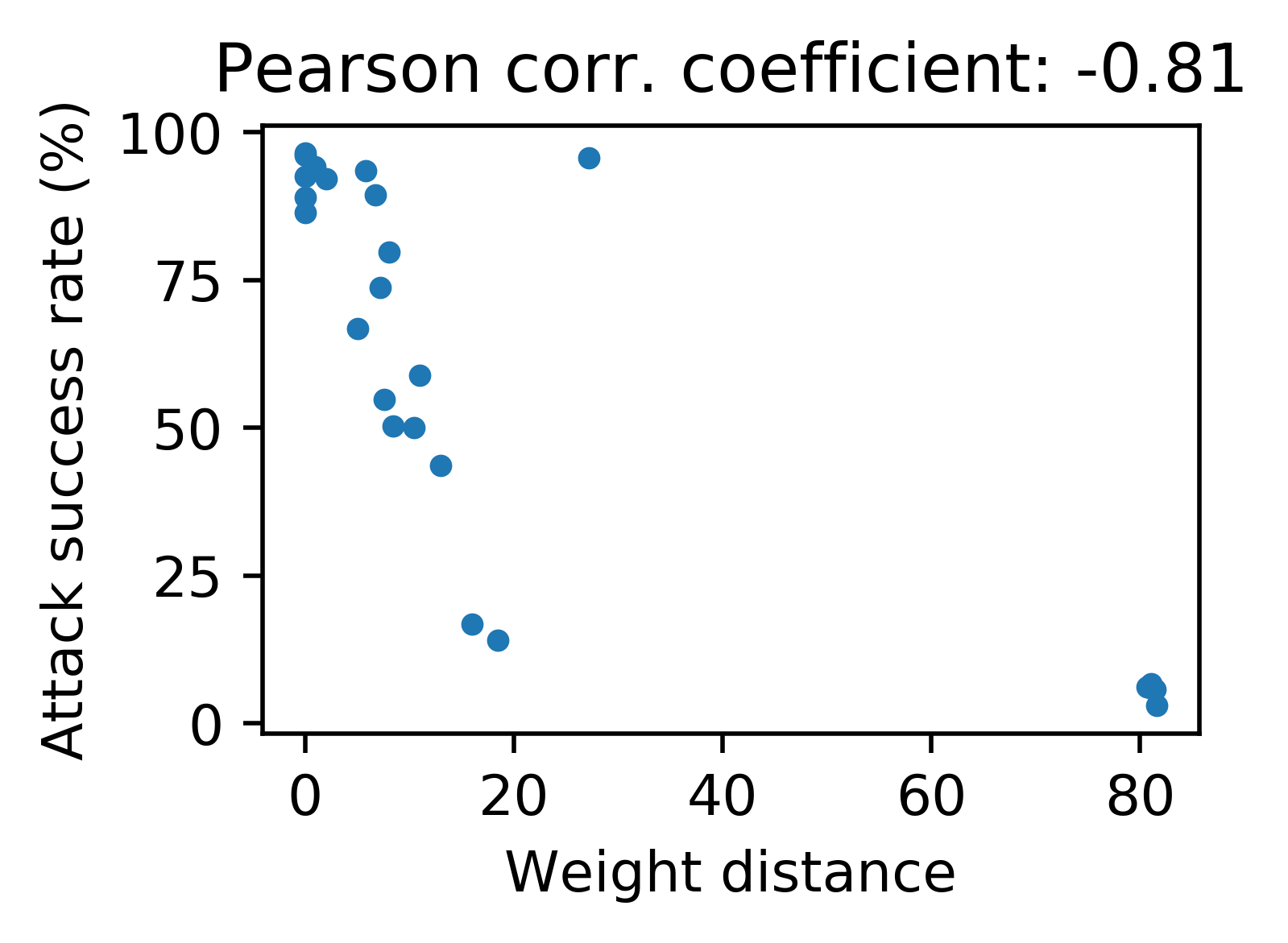}
    ~ 
    \includegraphics[width=0.48\linewidth]{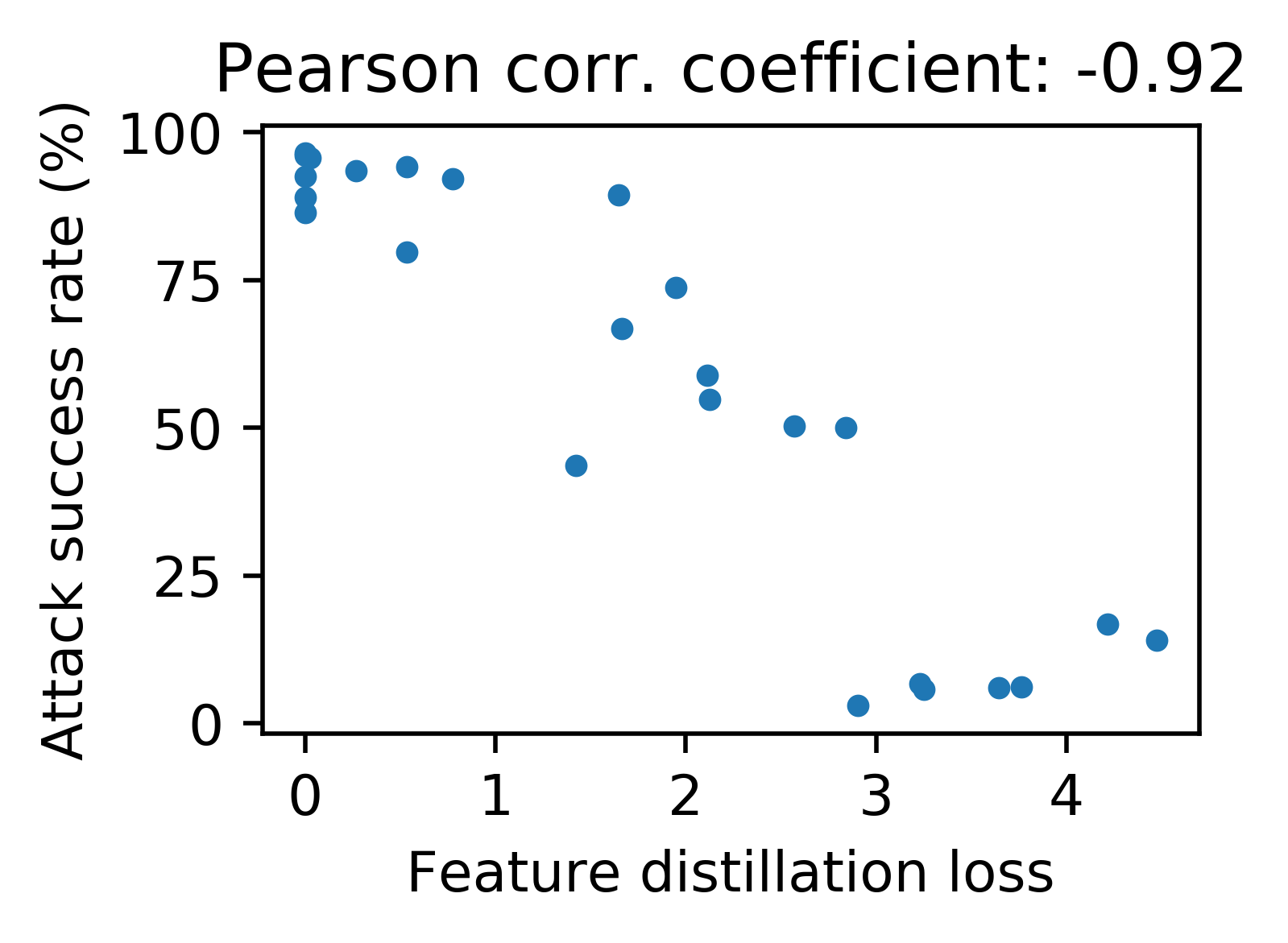}
    \caption{Robustness \textit{vs.} distance between transferred and pre-trained models for the five baseline methods on five datasets discussed in Table~\ref{table:benchmark}}
    \label{fig:robustness_corr}
\end{figure}

We conjecture that it is because the re-trained model has low similarity compared to the model under attack, \textit{i.e.}, the pre-trained model, while the fine-tuned models are initialized with the pre-trained weights that lead to potential similarity. To verify our conjecture, we measure the correlation between the attack success rate and the $\ell_2$ distance between the pre-trained and the transferred models. As for the distance measure, we look into two metrics. One is on the weight space, which measure the $\ell_2$ distance between the pre-trained weights and the transferred weights. The other metric is on the feature space, where we compute the $\ell_2$ feature distance averaged across different layers and training data (also known as the feature distillation loss). As shown in Figure~\ref{fig:robustness_corr}, the distance between the transferred and the pre-trained models negatively correlates with attack success rate for both distance measures, which matches our conjecture.

\section{Methodology}
\begin{figure}[h]
    \centering
    \includegraphics[width=0.8\linewidth]{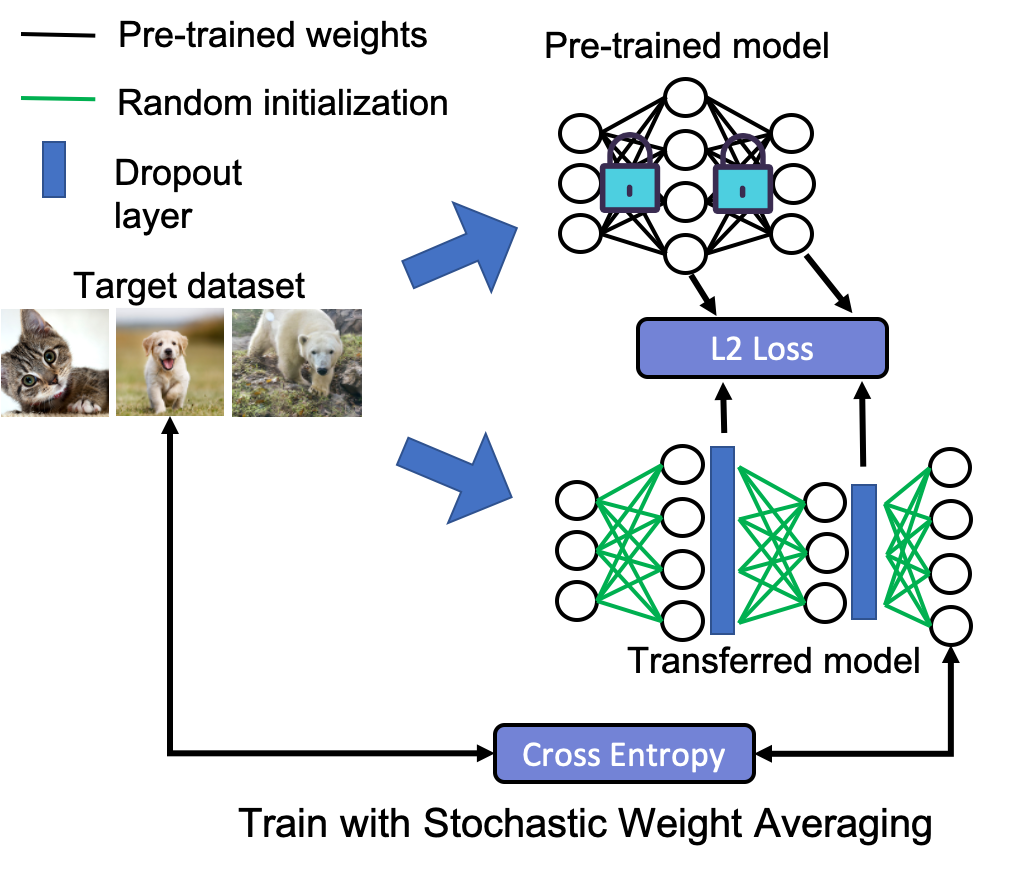}
    \caption{Schematic overview of Renofeation}
    \label{fig:scheme}
\end{figure}

To craft a defense mechanism based on our observation, the goal is to reduce model similarity without harming the benefits of transfer learning. To this end, we propose \underline{re}-training with \underline{no}isy \underline{fea}ture distilla\underline{tion}, or Renofeation for short. As shown in Fig.~\ref{fig:scheme}, Renofeation consists of two ingredients: (1) re-initialize the model with random initialization as opposed to inheriting the pre-trained weights and (2) train the model on the target dataset using noisy feature distillation. The first step removes the similarity with the pre-trained model that is embedded in the pre-trained weights. The second step uses feature distillation to encourage representation similarity to the pre-trained model for improving clean data performance while using a noisy process to discourage over-fitting to the pre-trained representation. As for the implementation for ``noisy" feature distillation, we adopt two regularization methods: spatial dropout~\cite{tompson2015efficient} during training and stochastic weight averaging~\cite{izmailov2018averaging}. 

\paragraph{Dropout} Dropout was proposed to avoid co-adaptation among neurons by randomly dropping out features during training~\cite{hinton2012improving}. In this work, we consider spatial-dropout~\cite{tompson2015efficient} for convolutional layers. Dropout has been used as a successful defense during evaluation time~\cite{wang2018defensive} which suggests that the adversarial features are highly co-adapted. As a result, instead of plain feature distillation that tries to mimic all the features of the pre-trained network, we propose to match the randomly dropped features to reduce the possibility of learning vulnerable features. We note that we do not randomly drop features during the evaluation time.

\paragraph{Stochastic Weight Averaging (SWA)} SWA has shown great promise in improving the generalization performance for deep neural networks~\cite{izmailov2018averaging}. The core idea is to average numerous local optima to form the final solution. It has been demonstrated empirically that SWA improves generalization while increasing the training loss. The rationale behind adopting SWA is that SWA is shown to be a successful technique that trades training loss for testing loss, which is exactly our goal: increasing the feature distillation loss without hurting the prediction performance. 

\section{Experiments}\label{sec:exp}
\subsection{Datasets and implementation detail}\label{sec:detail}
In this work, we consider five datasets to transfer to and models trained on ImageNet as pre-trained models. The datasets under consideration are shown in Table~\ref{table:dataset}.

\begin{table*}[t]
\caption{The characteristics of the datasets for transfer learning we considered in this work. We includes the number of training samples per class, the number of testing samples per class, and the number of classes.}
\begin{center}
\begin{small}
\begin{sc}
\begin{adjustbox}{max width=1\linewidth}
\begin{tabular}{c|c|c|c|c|c}
\toprule
Dataset & Task Category & \# Training Samples & \# Testing Samples & \# Classes & Abbreviation\\
\midrule
Stanford Dogs~\cite{khosla2011novel} & Fine-grained classification & 100 & $\approx$72 & 120 & Dog\\
Caltech-UCSD Birds~\cite{WelinderEtal2010} & Fine-grained classification & $\approx$30 & $\approx$ 29 & 200 & Bird\\
Stanford 40 Actions~\cite{yao2011human} & Action classification & 100 & $\approx$ 138& 40 & Action\\
MIT Indoor Scenes~\cite{quattoni2009recognizing} & Indoor scene classification & 80 & 20 & 67 & Indoor\\
102 Category Flower~\cite{nilsback2008automated} & Fine-grained classification & 20 & $\approx$60 & 102 & Flower\\
\bottomrule
\end{tabular}\label{table:dataset}
\end{adjustbox}
\end{sc}
\end{small}
\end{center}
\end{table*}

For training, we use a batch size of 64 and stochastic gradient descent with momentum following prior art~\cite{li2018delta,xuhong2018explicit}. For the experiments using fine-tuning, \textit{i.e.}, those that start with pre-trained weights, we use 30,000 iterations to make sure the loss converges. Additionally, we tune the learning rate, weight decay, and momentum for fine-tuning each dataset according to prior art~\cite{Li2020Rethinking}. Specifically, we tune learning rate$~\in\{0.01, 0.005\}$, momentum$~\in\{0, 0.9\}$, and weight decay$~\in\{0, 10^{-4}\}$ using grid search. For re-training, the hyper-parameters are set throughout the experiments across datasets without tuning. We use 90,000 iterations, learning rate $0.01$, momentum 0.9, and weight decay 0.005. Also, weight decay for the last linear layer is set to 0.01 across all the experiments following~\cite{li2018delta,xuhong2018explicit}. We use cosine learning rate decay for all the experiments. For fine-tuning methods that come with hyper-parameters such as L2SP and DELTA, we tune $\lambda$ to obtain the best transferred results according to prior art~\cite{xuhong2018explicit,li2018delta}.

We apply SWA by training with half of the learning rate, \textit{i.e.}, 0.005, as suggested in prior art~\cite{izmailov2018averaging}. SWA training considered has constant learning rate with 30,000 iterations. We average the models every 500 iterations. We insert the dropout layer after those that are used for the feature distillation loss and we use a dropout rate of 10\%. Regarding the parameters for crafting the adversarial examples, we set the perturbation budget $B$ to 0.4, the number of iterations of PGD to be 40, $m$ to be 1000 (following~\cite{Rezaei2020A}), and the learning rate to be 0.01. We set the target $t$ to be one-hot that always have one in the first neuron and zero for other neurons. We use AdverTorch~\cite{ding2019advertorch} for generating adversarial examples using the above specified objective and parameters.

\subsection{Ablating the proposed components}\label{sec:ablation}
In this subsection, we are interested in understanding the importance of the different components in the proposed Renofeation. Specifically, we would like to understand the impact of random initialization, dropout, and stochastic weight averaging. To do so, we have three baselines: (1) DELTA, which is the best transfer learning method in clean performance as shown in Table~\ref{table:benchmark}, (2) Re-training without transfer, which is the most robust method in Table~\ref{table:benchmark}, and (3) DELTA-R, which is DELTA with random (as opposed to pre-trained weights) initialization. For each of these baselines, we add dropout (DO), stochastic weight average (SWA), and both of them to see how these techniques affect the clean data performance and attack success rate. We conduct all the experiments in this subsection using ResNet-18. 

\begin{figure*}[t!]
    \centering
    \includegraphics[width=0.48\linewidth]{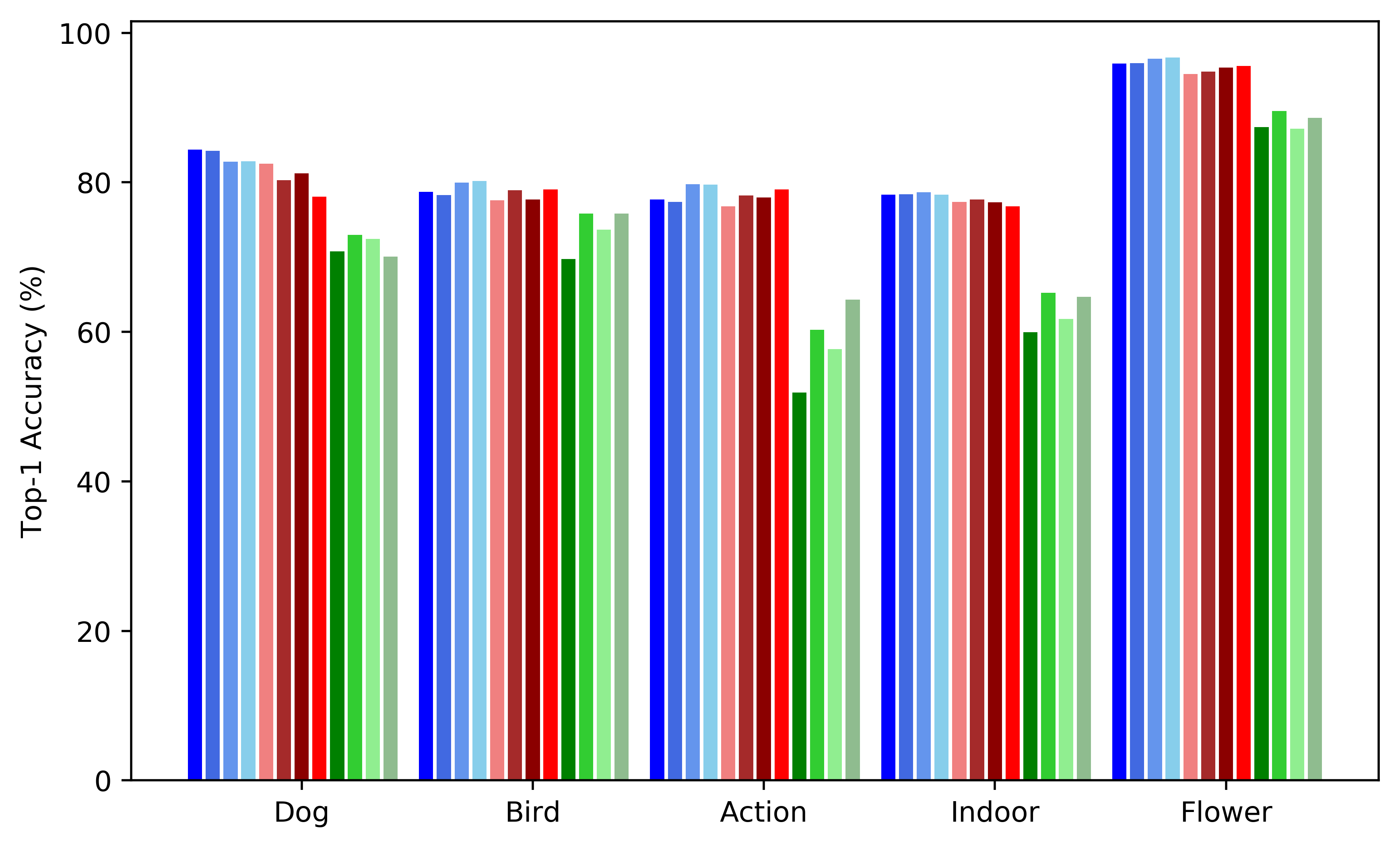}
    ~ 
    \includegraphics[width=0.48\linewidth]{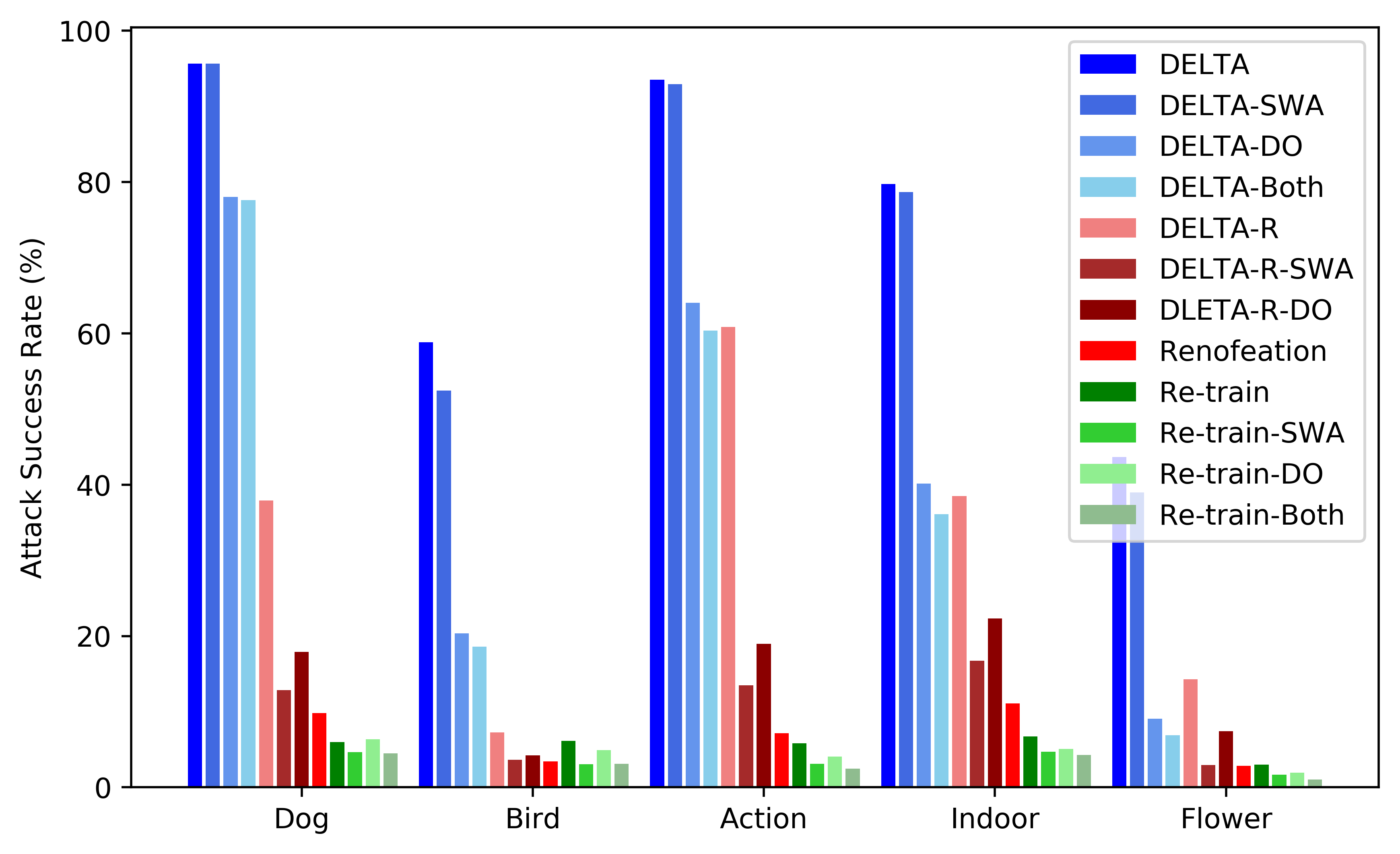}
    \caption{Ablation study of the effect of dropout (DO) and SWA on (Left) clean-data performance and (Right) attack success rate for the two baselines including DELTA and re-training.}
    \label{fig:ablation}
\end{figure*}

\paragraph{Importance of random initialization} From Table~\ref{table:benchmark}, we can observe that DELTA has the best clean data performance while re-training has the best robustness. Since DELTA achieves transfer via pre-trained weights and feature distillation, an interesting question arises: \textit{Do pre-trained weights help clean data performance and hurt robustness equally?} To answer this question, we compare DELTA and DELTA with random initialization (DELTA-R) in both clean data performance and ASR. As shown in Figure~\ref{fig:ablation}, we find that the pre-trained weights merely help clean data performance at the presence of feature distillation but hurts robustness significantly. This suggests that random initialization effectively makes the function less similar to the function induced by the pre-trained weights even in the presence of feature distillation. This is plausible because feature distillation is enforcing the two functions to be similar only at the training data points, which are scarce for transfer learning.

\paragraph{The effect of regularization} In Renofeation, both dropout (DO) and stochastic weight averaging (SWA) are proposed to be incorporated into the transfer learning process. It would be of interest to understand their respective impact on both robustness and clean data performance for all three baselines. Let us first focus on DELTA, as shown in Fig.~\ref{fig:ablation}, we can observe that SWA does not work well in improving the robustness for DELTA, which might be due to the local optimality of the pre-trained weights that leads to less diverse model weights throughout the fine-tuning process. On the other hand, DO improves the robustness for DELTA, but the robustness still falls short when compared to re-training. Both techniques marginally improve the clean data performance for DELTA except for the Dog dataset, where we see a slight accuracy drop.

Considering SWA and DO for re-training, we find that both techniques significantly improve the clean data performance for re-training and this is expected based on the results from the DO and SWA papers. Nonetheless, the clean data performance of Re-train with both techniques still underperforms DELTA by a significant margin in most datasets.

Lastly, we consider SWA and DO for DELTA-R. We find that SWA works much better for improving robustness when compared to applying SWA to DELTA. This suggests that the weight initialization greatly affects the training trajectory. When considering the clean data performance, both techniques again marginally improve the clean data performance for DELTA-R except for the Dog dataset. Overall, Renofeation, which consists of DELTA-R, DO, and SWA, achieves the best of both worlds, with clean data performance comparable to DELTA and the robustness comparable to re-training.

\begin{table*}[t]
\caption{Comparing DELTA, Renofeation, and re-training for different ConvNets. Renofeation has the clean data performance comparable to DELTA and robustness comparable to re-training for different ConvNets we study.}
\vskip 0.15in
\begin{center}
\begin{small}
\begin{sc}
\begin{adjustbox}{max width=1\linewidth}
\begin{tabular}{c|c|c|c|c|c|c|c||c}
\toprule
\multicolumn{3}{c|}{} & Dog & Bird & Action & Indoor & Flower & Average\\
\midrule
\multirow{6}{*}{ResNet-18} & \multirow{2}{*}{DELTA} & clean & \textbf{84.39} & 78.75 & 77.69 & \textbf{78.36} & \textbf{95.90} & -\\
& & ASR & 95.65 & 58.83 & 93.51 & 79.71 & 43.65 & 74.27\\
\cline{2-9}
 & \multirow{2}{*}{Renofeation} & clean & 78.11 & \textbf{79.03} & \textbf{79.07} & 76.79 & 95.59 & -\\
& & ASR & 9.83 & \textbf{3.41} & 7.16 & 11.08 & \textbf{2.86} & 6.87\\
\cline{2-9}
& \multirow{2}{*}{Re-training} & clean & 70.77 & 69.76 & 51.90 & 59.93 & 87.38 & -\\
& & ASR & \textbf{5.99} & 6.14 & \textbf{5.82} & \textbf{6.73} & 3.00 & 5.54\\
\midrule
\midrule
\multirow{6}{*}{ResNet-50} & \multirow{2}{*}{DELTA} & clean & \textbf{90.13} & \textbf{81.95} & 81.87 & 79.93 & 96.63 & -\\
& & ASR & 94.69 & 32.29 & 91.94 & 84.69 & 24.84 & 65.69\\
\cline{2-9}
 & \multirow{2}{*}{Renofeation} & clean & 83.57 & 79.27 & \textbf{84.04} & \textbf{80.67} & \textbf{96.75} & -\\
& & ASR & \textbf{5.08} & \textbf{3.96} & \textbf{3.33} & 7.12 & 2.39 & 4.38\\
\cline{2-9}
& \multirow{2}{*}{Re-training} & clean & 72.55 & 70.47 & 53.53 & 59.11 & 85.93 & -\\
& & ASR & 6.30 & 7.45 & 6.11 & \textbf{6.06} & \textbf{2.20}& 5.62\\
\midrule
\midrule
\multirow{6}{*}{ResNet-101} & \multirow{2}{*}{DELTA} & clean & \textbf{91.92} & \textbf{82.07} & 82.61 & 80.00 & \textbf{96.37} & -\\
& & ASR & 88.03 & 42.60 & 87.53 & 89.27 & 44.04 & 70.29\\
\cline{2-9}
 & \multirow{2}{*}{Renofeation} & clean & 83.88 & 80.98 & \textbf{84.67} & \textbf{80.97} & 96.33 & -\\
& & ASR & \textbf{4.38} & \textbf{3.54} & \textbf{3.69} & 9.95 & 3.09 & 4.93\\
\cline{2-9}
& \multirow{2}{*}{Re-training} & clean & 73.42 & 71.80 & 52.78 & 61.12 & 85.59 & -\\
& & ASR & 6.64 & 7.21 & 6.64 & \textbf{5.13} & \textbf{2.00} & 5.52\\
\midrule
\midrule
\multirow{6}{*}{MobileNetV2} & \multirow{2}{*}{DELTA} & clean & \textbf{84.86} & \textbf{78.51} & \textbf{78.94} & 76.12 & \textbf{96.68} & -\\
& & ASR & 82.89 & 40.30 & 57.00 & 52.45 & 21.08 & 50.75\\
\cline{2-9}
 & \multirow{2}{*}{Renofeation} & clean & 76.42 & 75.70 & 77.78 & \textbf{76.49} & 96.32 & -\\
& & ASR & 11.62 & \textbf{6.79} & \textbf{5.92} & \textbf{7.12} & 2.84 & 6.86\\
\cline{2-9}
& \multirow{2}{*}{Re-training} & clean & 67.95 & 69.50 & 52.86 & 61.49 & 88.73 & -\\
& & ASR & \textbf{8.56} & 8.54 & 8.35 & 7.65 & \textbf{2.71} & 7.16\\
\bottomrule
\end{tabular}\label{table:cnns}
\end{adjustbox}
\end{sc}
\end{small}
\end{center}
\vskip -0.1in
\end{table*}

\subsection{More networks}
So far, we have conducted our experiments and analyses based on ResNet-18. We are interested to see if Renofeation is still more preferable compared to re-training and DELTA for other networks. Specifically, we further consider deeper networks, \textit{i.e.}, ResNet-50, and ResNet-101. Additionally, due to recent interests in reducing the computational overhead of ConvNets for deployment purpose~\cite{wang2020pay,stamoulis2019single,sandler2018mobilenetv2,chin2019legr,wu2019fbnet}, we also consider a compact network, \textit{i.e.}, MobileNetV2~\cite{sandler2018mobilenetv2}. Due to computational considerations, for DELTA with other networks, we inherit the learning rate, weight decay, and momentum from ResNet-18 for each of the dataset. 

As shown in Table~\ref{table:cnns}, Renofeation achieves clean-data performance comparable to that of DELTA and has robustness similar to re-training across all ConvNets we have investigated. We note that while Renofeation in general has clean data performance comparable to DELTA, it is not the case for the Dog dataset, where DELTA consistently has higher accuracy compared to Renofeation. We hypothesize that the accuracy loss in this case may be due to the Dog dataset being a strict subset of the ImageNet dataset and therefore matching features alone on the scarce target dataset may not be sufficient to recover the features for good generalization. This can be inferred from the fact that the linear classifier alone has clean data performance matching DELTA for the Dog dataset as shown in Table~\ref{table:benchmark}. While it is less likely to conduct transfer learning to a target dataset that is a strict subset of the source dataset, this phenomenon also suggests that there is room for improvement for future research to tackle the studied threat model.

\subsection{Data amount ablation}
From previous results, we show that feature distillation using the target dataset is able to achieve competitive clean data performance compared to fine-tuning. Intuitively, if the amount of training data is large, feature distillation should be able recover the knowledge encoded in the pre-trained weights. However, in the transfer learning case, target datasets usually have much less training data compared to large-scale datasets such as ImageNet. In this section, we ablate the number of training samples to understand how it affects the effectiveness of Renofeation so as to further provide a guideline for when to use it. Specifically, we consider cases where the training data for each dataset is reduced to $33\%$ and $66\%$. For each class in the dataset, we randomly sub-sample $33\%$ and $66\%$ of the training images. As a result, the overall training dataset is still balanced across classes. 

As shown in Table~\ref{table:sample_counts}, we find that as the training data size decreases, the clean data performance gap between Renofeation and DELTA increases. This is expected as feature distillation with fewer samples makes it an underdetermined problem to match the function of the pre-trained model. However, Renofeation still greatly improves over DELTA in robustness and greatly improves over re-training in clean data performance. 

\begin{table}[h]
\caption{Ablating the number of training samples for each dataset to 33\% and 66\% and compare the performances among methods. As the training data gets smaller in size, Renofeation provides more improvement in clean data performance compared to re-training while having robustness much better than DELTA. }
\begin{center}
\begin{small}
\begin{sc}
\begin{adjustbox}{max width=1\linewidth}
\begin{tabular}{c|c|c|c|c|c|c|c}
\toprule
\multicolumn{3}{c|}{} & Dog & Bird & Action & Indoor & Flower\\
\midrule
\multirow{6}{*}{33\%} & \multirow{2}{*}{DELTA} & clean & \textbf{81.80} & \textbf{63.41} & \textbf{70.72} & \textbf{70.97} & \textbf{90.11}\\
& & ASR & 95.77 & 74.12 & 93.94 & 85.38 & 46.60\\
\cline{2-8}
 & \multirow{2}{*}{Renofeation} & clean & 74.13 & 61.75 & 69.22 & 70.22 & 88.32\\
& & ASR & 10.88 & \textbf{5.56} & 9.40 & 15.73 & 4.94\\
\cline{2-8}
& \multirow{2}{*}{Re-training} & clean & 44.98 & 26.10 & 24.51 & 37.54 & 62.73\\
& & ASR & \textbf{9.67} & 14.68 & \textbf{9.22} & \textbf{8.35} & \textbf{2.85}\\
\midrule
\multirow{6}{*}{66\%} & \multirow{2}{*}{DELTA} & clean & \textbf{83.58} & 73.04 & 75.52 & \textbf{75.30} & \textbf{94.23}\\
& & ASR & 95.36 & 64.58 & 93.80 & 80.77 & 56.39\\
\cline{2-8}
 & \multirow{2}{*}{Renofeation} & clean & 77.25 & \textbf{74.46} & \textbf{76.09} & 74.48 & 93.56\\
& & ASR & 9.85 & \textbf{3.55} & 7.53 & 12.22 & 4.73\\
\cline{2-8}
& \multirow{2}{*}{Re-training} & clean & 64.03 & 56.47 & 40.73 & 52.61 & 80.60\\
& & ASR & \textbf{7.72} & 10.79 & \textbf{5.86} & \textbf{7.23} & \textbf{3.23}\\
\bottomrule
\end{tabular}\label{table:sample_counts}
\end{adjustbox}
\end{sc}
\end{small}
\end{center}
\end{table}

\subsection{Adversarial training}
\begin{table}[h]
\caption{Comparison among DELTA, DELTA with PGD-3 adversarial training, and proposed Renofeation. Renofeation has the best robustness with comparable clean data performance with other DELTA variants.}
\begin{center}
\begin{small}
\begin{sc}
\begin{adjustbox}{max width=\linewidth}
\begin{tabular}{c|c|c|c|c|c|c}
\toprule
\multicolumn{2}{c|}{} & Dog & Bird & Action & Indoor & Flower\\
\midrule
\multirow{2}{*}{DELTA} & clean & \textbf{84.39} & 78.75 & 77.69 & \textbf{78.36} & 95.90\\
& ASR & 95.65 & 58.83 & 93.51 & 79.71 & 43.65\\
\midrule
\multirow{2}{*}{DELTA Adv. Trained} & clean & 82.83 & 77.10 & 75.69 & 77.84 & 95.12\\
& ASR & 85.86 & 16.77 & 85.19 & 61.84 & 23.85\\
\midrule
\multirow{2}{*}{DELTA Adv. Trained + SWA + DO} & clean & 81.42 & \textbf{80.20} & \textbf{79.12} & 78.28 & \textbf{96.81}\\
& ASR & 53.03 & 8.93 & 63.72 & 36.60 & 4.92\\
\midrule
\multirow{2}{*}{Renofeation} & clean & 78.11 & 79.03 & 79.07 & 76.79 & 95.59\\
& ASR & \textbf{9.83} & \textbf{3.41} & \textbf{7.16} & \textbf{11.08} & \textbf{2.86}\\
\bottomrule
\end{tabular}\label{table:advtrain}
\end{adjustbox}
\end{sc}
\end{small}
\end{center}
\end{table}

While we showed that our proposed Renofeation approach, when compared to DELTA, achieves better robustness with comparable clean-data performance under our threat model, adversarial training can also be considered as a defense under our threat model. As a result, in this section, we compare our method with adversarial training to further demonstrate its effectiveness. To conduct adversarial training in our considered threat model, we train DELTA with 2$\times$ more iterations and, we craft adversarial examples with three iterations of projected gradient descent. As shown in Table~\ref{table:advtrain}, adversarial training indeed achieves better robustness compared to the baselines but worse compared to Renofeation. This is because, as a defense method, Renofeation has used the prior that the attack is generated using the pre-trained model and defends accordingly using random initialization while adversarial training has not harnessed this prior.

\subsection{Regularization and feature distillation}
We have shown in previous sections that both dropout (DO) and stochastic weight averaging (SWA) are helpful in reducing the attack success rate. It is not clear if this happens due to the reasons we expected: ``reduce overfitting in terms of the feature distillation loss." As a result, we analyze the impact on feature distillation loss when DELTA-R is augmented with DO and SWA. As shown in Table~\ref{table:tricks}, we observe that these regularization techniques indeed increase the feature distillation loss, which in turn improves the robustness of DELTA-R. Additionally, we find empirically that both dropout and SWA can work together to achieve better regularization.

\begin{table}[h]
\caption{The effect of dropout and SWA on feature distillation loss, clean-data performance, and robustness for DELTA-R and ResNet-18.}
\vskip 0.15in
\begin{center}
\begin{small}
\begin{sc}
\begin{adjustbox}{max width=1\linewidth}
\begin{tabular}{c|c|c|c|c|c|c}
\toprule
\multicolumn{2}{c|}{} & Dog & Bird & Action & Indoor & Flower\\
\midrule
\multirow{3}{*}{DELTA-R} & clean & \textbf{82.49} & 77.58 & 76.79 & \textbf{77.39} & 94.49\\
& ASR & 37.93 & 7.28 & 60.83 & 38.48 & 14.30\\
& Feature loss & 0.70 & 1.48 & 0.72 & 0.68 & 0.56\\
\midrule
\multirow{3}{*}{DELTA-R + Dropout} & clean & 81.21 & 77.72 & 78.00 & 77.31 & 95.35\\
& ASR & 17.91 & 4.24 & 18.98 & 22.30 & 7.44\\
& Feature loss & 0.86 & 1.57 & 1.03 & 0.89 & 0.68\\
\midrule
\multirow{3}{*}{DELTA-R + SWA} & clean & 80.32 & 78.92 & 78.07 & 77.69 & 94.81\\
& ASR & 12.87 & 3.65 & 23.62 & 16.72 & 2.95\\
& Feature loss & 0.86 & 1.63 & 0.87 & 0.82 & 0.73\\
\midrule
\multirow{3}{*}{Renofeation} & clean & 78.11 & \textbf{79.03} & \textbf{79.07} & 76.79 & \textbf{95.59}\\
& ASR & \textbf{9.83} & \textbf{3.41} & \textbf{7.16} & \textbf{11.08} & \textbf{2.86}\\
& Feature loss & \textbf{1.00} & \textbf{1.68} & \textbf{1.06} & \textbf{1.02} & \textbf{0.81}\\
\bottomrule
\end{tabular}\label{table:tricks}
\end{adjustbox}
\end{sc}
\end{small}
\end{center}
\vskip -0.1in
\end{table}

\subsection{Tuning hyperparameters for DELTA}
\begin{figure}[h]
    \centering
    \includegraphics[width=0.8\linewidth]{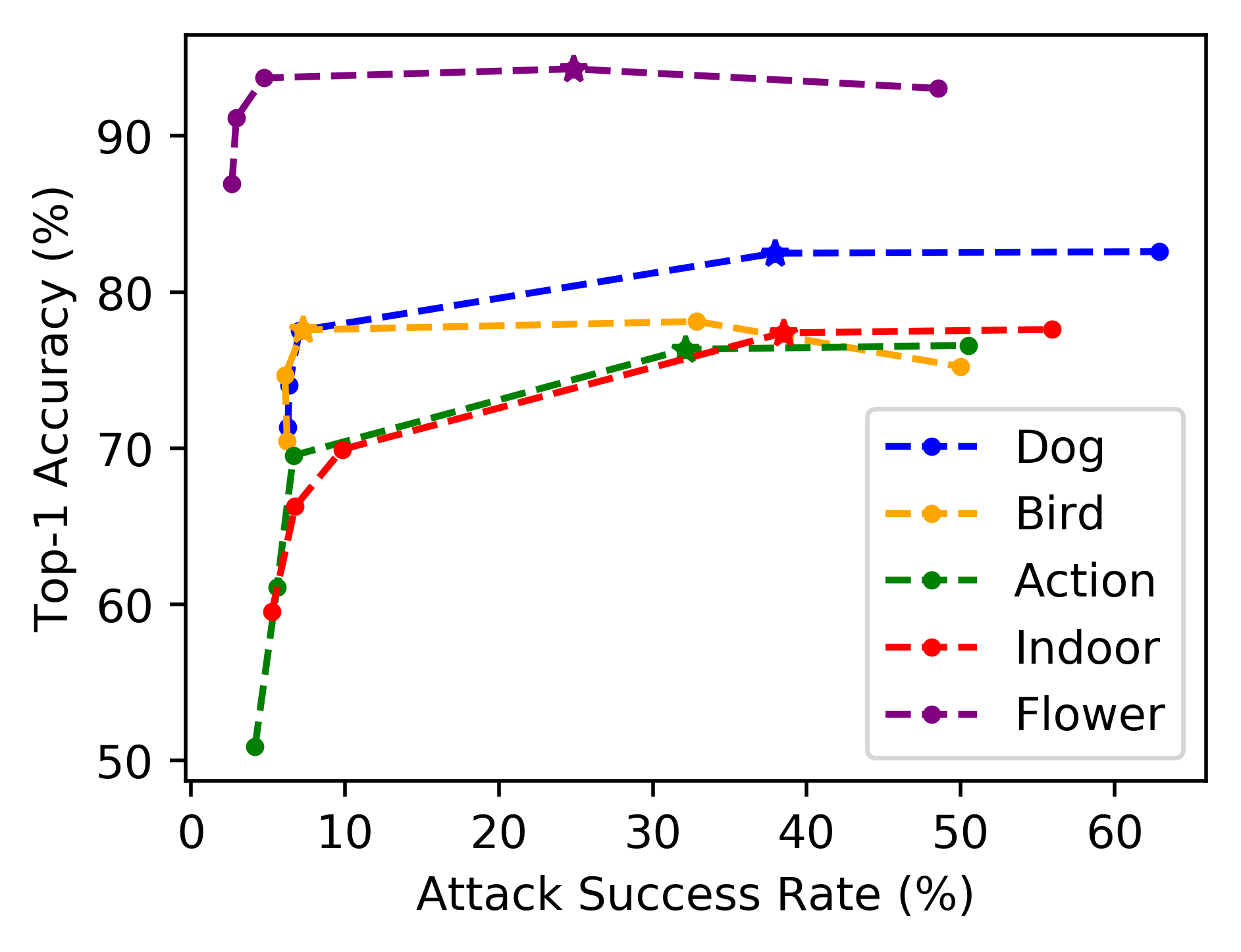}
    \caption{The effect of tuning $\lambda_{feat}$ on the trade-off between clean-data performance and the attack success rate for ResNet-18. Star marks the $\lambda_{feat}$ we use.}
    \label{fig:lambda}
\end{figure}

A natural idea to reduce the impact of feature distillation is to tune its corresponding weight ($\lambda_{feat}$) on the training loss. As shown in Figure~\ref{fig:lambda}, even the best $\lambda_{feat}$ still incurs high ASR for datasets such as Indoor, Dog, and Action. The performance gained obtained by Renofeation cannot be obtained by simply tuning the weight for the feature distillation loss.

\section{Conclusion}
In this work, we first show that the attack proposed by Rezaei \textit{et al.}~\cite{Rezaei2020A} works not only for transfer learning by re-training the last linear layer, but also for end-to-end fine-tuning. This is concerning due to the widely adopted fine-tuning paradigm. We show that the attack success rate correlates well with the similarity between the pre-trained and the fine-tuned model. Based on this observation, we propose Renofeation, a transfer learning method that is significantly more robust to adversarial attacks crafted based on the pre-trained model when compared to state-of-the-art transfer learning methods based on fine-tuning. Renofeation has two key ingredients: (1) random initialization and (2) noisy feature distillation. We have extensively analyzed the proposed Renofeation empirically with ablation to demonstrate its effectiveness. While the threat model under consideration is relatively new~\cite{Rezaei2020A}, it is crucial to improve robustness under this threat model due to the practical popularity of fine-tuning. This work takes a first step towards improving the robustness under this threat model and sheds light on this topic for future study. 

\section*{Acknowledgement}
This research was supported in part by NSF CCF Grant No. 1815899, NSF CSR Grant No. 1815780, and NSF ACI Grant No. 1445606 at the Pittsburgh Supercomputing Center (PSC).

{\small
\bibliographystyle{ieee_fullname}
\bibliography{egbib}
}

\end{document}